\documentclass{article}

\usepackage{microtype}
\usepackage{graphicx}
\usepackage{subfigure}
\usepackage{booktabs} 
\usepackage{multirow}
\usepackage{tabularx} %
\usepackage[hyphens]{url}
\usepackage{hyperref}

\usepackage[accepted]{icml2021}

\icmltitlerunning{Benchmarking Differential Privacy and Federated Learning for BERT Models}

\begin{document}

\twocolumn[
\icmltitle{Benchmarking Differential Privacy and Federated Learning for BERT Models}



\icmlsetsymbol{equal}{*}

\begin{icmlauthorlist}
\icmlauthor{Priyam Basu}{equal,man}
\icmlauthor{Tiasa Singha Roy}{equal,man}
\icmlauthor{Rakshit Naidu}{man,cmu,om}
\icmlauthor{Zumrut Muftuoglu}{om,ytu}
\icmlauthor{Sahib Singh}{om,ford}
\icmlauthor{Fatemehsadat Mireshghallah}{ucsd,om}
\end{icmlauthorlist}

\icmlaffiliation{man}{Manipal Institute of Technology}
\icmlaffiliation{cmu}{Carnegie Mellon University}
\icmlaffiliation{om}{OpenMined}
\icmlaffiliation{ytu}{Yildiz Technical University}
\icmlaffiliation{ford}{Ford Motor Company}
\icmlaffiliation{ucsd}{University of California, San Diego}

\icmlcorrespondingauthor{Rakshit Naidu}{rnemakal@andrew.cmu.edu}

\icmlkeywords{Machine Learning, ICML}

\vskip 0.3in
]



\printAffiliationsAndNotice{\icmlEqualContribution} 
\newcommand{\FM}[1]{\textcolor{blue}{\bf \small [ #1 --Fatemeh]}}

\begin{abstract}
 Natural Language Processing  (NLP) techniques can be applied to help with the diagnosis of medical conditions such as depression, using a collection of a person's utterances.
 %
 %
 Due to the sensitive nature of such data, privacy measures need to be taken for handling and training models with such data. 
 In this work we study the effects that the application of Differential Privacy (DP) has, in both a centralized and a Federated Learning (FL) setup, on training contextualized language models (BERT, ALBERT, RoBERTa and DistilBERT).
 We offer insights on how to privately train NLP models and what architectures and setups provide more desirable privacy utility trade-offs.
 We envisage this work to be used in future  healthcare and mental health studies to keep medical history private. Therefore, we provide an open-source implementation of this work~\footnote{\url{https://github.com/whopriyam/Benchmarking-Differential-Privacy-and-Federated-Learning-for-BERT-Models}}. 
\end{abstract}

\vspace{-2ex}
\section{Introduction}
Mental health is defined as a “state of well-being in which individuals realize their potential, cope with the normal stresses of life, work productively, and contribute to their communities” by the World Health Organization (WHO)~\citet{mhactionplan13}. Depression is a very common mental disease that a large number of people throughout the world suffer from. According to a study conducted by WHO, in 2020 more than 264 million people all over the world suffer from depression~\citet{depressionwho20}.
Given the status quo, technology can offer new ways of diagnosing depression, and it can also help in treatments. One of these methods is using linguistic markers, as many people express their feelings and thoughts on social media, or in personal journals and well-being applications. Social media platforms let us observe the activities, thoughts, and feelings of people’s daily lives, including those of patients suffering from mental disorders~\cite{depspanishtweets19}. Through analyzing these markers, researchers can build models to help with the diagnosis of such conditions.
%

Linguistic markers in Tweets or other forms of utterance can be used to create statistical models that can detect and predict depression, and some other forms of mental illnesses. However, due to the sensitive nature of such data~\cite{youarewhatyoutweet19}, training models on them and realising them can have major consequences. 
%
%
Public Tweets do not necessarily raise privacy concerns, however, Tweets on sensitive subjects should be treated with more care, especially since the account user can make their account private at any point (making previously collected public tweets now technically private).
%
%
Literature shows that privacy-enhancing approaches (such as differential privacy, federated learning and homomorphic encryption) are promising approaches in handling these concerns. However, prior work~\cite{smith2018federated, singh2020benchmarking, li-etal-2018-towards} doesn't study the compound effects of differential privacy and federated learning on large NLP models, such as BERT.


The goal of this paper is to
benchmark the effect of privacy measures such as differential privacy, on the utility of central and federated training of BERT-based models. We explore different privacy budgets, $\varepsilon$ and observe how they affect the utility of models trained on depression and sexual harassment-related Tweets. For the federated setup, we explore both the IID and non-IID distributions of data.
Our empirical studies provide insights into which model architectures and privacy regimes provide  more desirable privacy-utility trade-offs, and what are the next steps in the direction of federating NLP models on privacy-protected data. 
To facilitate research in this direction, we have made our framework public available in this Github repository:~\href{https://github.com/whopriyam/Benchmarking-Differential-Privacy-and-Federated-Learning-for-BERT-Models}{Benchmarking DP and FL for BERT models}.

\begin{figure}
\includegraphics[width=\linewidth]{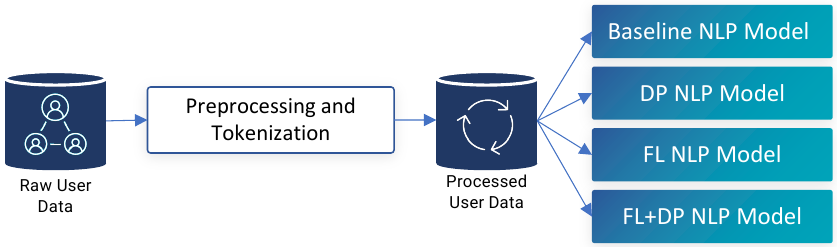}
\vspace{-5ex}
\caption{Pipeline of our benchmarking framework: We preprocess raw Twitter data, and then use  it to run four sets of experiments comparing  conventional training in a centralized setup, training with differential privacy in a centralized setup, training  with federated learning in a distributed setup and finally, applying differential privacy to federated learning. 
}
\vspace{-5ex}
\label{fig:Figure1}
\end{figure}


\section{Related Work}
%
Differential privacy~\cite{Dwork2011, Dwork06} which will be explored further in the next section, uses random noise to provide that the publicly visible information doesn’t change much if one individual in the dataset is removed. On the other hand, deep learning  techniques are used to earn text representation via neural models by language application~\cite{Bengio03, Mikolov13, devlin-etal-2019-bert}. And mostly, the input text gives individual information about the author, such as demographic data. Some sentiment analysis results show that user attributes can be detected easily~\cite{Dirk15, Rosso2019}.There have been studies training differentially private deep models with the formal differential privacy approach in the literature \cite{Abadi_2016, mcmahan2018learning, Yu_2019}. ~\citeauthor{preotiuc-pietro-etal-2015-analysis} showed that demographic information about the text's author can be predicted through linguistic cues in the text. To tackle this kind of privacy concerns,~\citet{naacl-2021-textgen} and ~\citet{li-etal-2018-towards} propose training models using adversarial learning to improve the robustness and privacy of neural representation in natural language generation and classification  tasks, respectively.

%
Federated learning is another privacy-enhancing approach~\cite{mcmahan2017communicationefficient, yang2019federated, kairouz2021advances, jana2021investigation}, which relies on distributed training of models on devices, and sharing of gradients. 
%
%
~\citeauthor{Sarma21} show that it is possible to train data  multi-institutionally without centralizing or sharing the underlying physical data through federated learning. There have been studies focusing on coping with the statistical challenges ~\cite{smith2018federated, Zhao18} and security concerns~\cite{Segal17, geyer2018differentially}. There also have been studies to customize federated learning~\cite{chen2019federated, smith2018federated}. Google proposed a horizontal federated learning solution for Android phone model updates~\cite{McMahan2016FederatedLO}.~\citeauthor{Segal17} introduce 
a secure aggregation scheme to guard aggregated user updates under their federated learning framework.
The work done by~\citeauthor{singh2020benchmarking} also conducts similar benchmarking experiments on healthcare data however their work is focused on image data and vision tasks.

\vspace{-1ex}
\section{Preliminaries}
\subsection{Natural Language Processing}
The field of Natural Language Processing (NLP) which is a sub-category of Artificial Intelligence and linguistics enables analysis and understanding or reproduction of the canonical structure of natural languages. Information extraction, machine translation, summarization, search and human-computer interfaces are the result of the NLP applications \cite{collobert08}. Although complete semantic understanding seems far-distant, there are promising applications in the literature. With the development of the social web, new content-sharing services allow people to create and share their own contents, ideas, and opinions, in a time- and cost-efficient way, with virtually millions of other people connected to the World Wide Web which means a massive amount of information. But since this massive amount of information is mainly unstructured, It can not be processed by machines directly \cite{cambria14}.

\noindent\textbf{BERT.} The NLP models that have been implemented in this paper are Transformer based models because they use self-attention mechanism and process the entire input data at once instead of as a sequence to capture long term dependencies for obtaining contextual meaning. BERT (Bidirectional Encoder Representations from Transformers) \cite{devlin-etal-2019-bert} masks 15\% of the input tokens (defined using Wordpiece) for the model to detect the masked word. [SEP] and [CLS] tokens are also added to separate between two sentences and classification for Next Sentence Prediction (NSP). segment embedding that identifies each sentence, and a position embedding to distinguish the position of each token and replace recurrence. The final layer of BERT contains a token representation $T_{i}$ and the classifier embedding $C$, then each $T_{i}$ is used to predict whether the token was masked or not and the $C$ representation to predict if the two sentences were contiguous or not.

\noindent\textbf{RoBERTa.} Robustly Optimized BERT-Pretraining Approach (RoBERTa) \cite{liu2019roberta} essentially includes fine-tuning the original BERT model along with data and inputs manipulation. To improve the training procedure, RoBERTa removes the Next Sentence Prediction (NSP) task from BERT’s pre-training and introduces static and dynamic masking so that the masked token changes during the training epochs. It uses 160 GB of text for pre-training, including 16GB of Books Corpus and English Wikipedia used in BERT. The additional data included CommonCrawl News dataset, Web text corpus and Stories from Common Crawl. For tokenization, RoBERTa uses a byte-level Byte-Pair Encoding (BPE) encoding scheme with a vocabulary containing 50000 subword units in contrast to BERT’s character-level BPE with a 30000 vocabulary. It is trained on larger batches without NSP objective in pre-training on larger sequences.

\noindent\textbf{DistilBERT.} Distilled version of BERT (DistilBERT) \cite{sanh2020distilbert} retains 97\% performance but uses only half as many parameters as BERT. It does not have token-type embeddings, pooler and retains only half of the layers from BERT. DistilBERT uses a technique called distillation, which approximates BERT, i.e. the large neural network by a smaller one. It follows the concept that once a large neural network has been trained, its full output distributions can be approximated using a smaller network. This is in some sense similar to posterior approximation. One of the key optimization functions used for posterior approximation in Bayesian Statistics is Kulback Leiber divergence and has naturally been used here as well. In Bayesian statistics, the true posterior is approximated whereas with distillation we are just approximating the posterior learned by the larger network.


\noindent\textbf{ALBERT.} A Lite BERT for Self-Supervised Learning of Language Representations (ALBERT) \cite{lan2020albert} uses Factorized embedding parameterization where the size of the hidden layers from the size of vocabulary embeddings is isolated by projecting one-hot vectors into a lower dimensional embedding space and then to the hidden space, which made it easier to increase the hidden layer size without significantly increasing the parameter size of the vocabulary embeddings. Cross-layer parameter sharing \cite{sachan2018parameter} is used for all parameters across layers to prevent the parameters from growing along with the depth of the network. As a result, the large ALBERT model has about 18 times fewer parameters compared to BERT-large. ALBERT also uses sentence-order prediction (SOP) loss to model inter-sentence coherence, which enables the new model to perform more robustly in multi-sentence encoding tasks.


\begin{table*}[h]
\centering
\footnotesize
  \caption{Average Test accuracies of models trained in centralized and FL setups, on the Depression dataset}
  \begin{tabularx}{{0.77\textwidth}}{@{}lrrrrr@{}}
  \toprule
   \textbf{Setup}	&\textbf{Epsilon}	&\textbf{BERT}	&\textbf{RoBERTa}	&\textbf{DistillBERT}	&\textbf{ALBERT}\\
   \midrule 
    \multirow{4}{*}{Centralized DP}	
    &0.5	&{53.20 $\pm$ 22.04} &{56.43 $\pm$ 24.13}	&{42.76 $\pm$ 24.29}	&{54.44 $\pm$ 22.48}\\ 
    &5	&{44.59 $\pm$ 23.26}	&{45.85 $\pm$ 15.23}	&{55.09 $\pm$ 22.76}	&{56.75 $\pm$ 12.42}\\ 
    &15	&{58.80 $\pm$ 19.09} &{39.76 $\pm$ 9.67} 	&{63.81 $\pm$ 10.86}	&{60.35 $\pm$ 11.58}\\ 
    &$\infty$ (No noise)	&{75.81 $\pm$ 0.00}	&{84.01 $\pm$ 0.00} &{74.47 $\pm$ 0.00}	&{64.52 $\pm$ 0.00}\\ 
   \midrule 
    \multirow{4}{*}{FL-IID}	
    &0.5	&{57.24 $\pm$ 23.28}	&{39.02 $\pm$ 35.95}	&{20.23 $\pm$ 20.43} &{58.10 $\pm$15.30}\\ 
    & 5	&{70.92 $\pm$ 0.84}	&{39.13 $\pm$ 35.78}	&{54.43 $\pm$ 18.54}	&{45.49 $\pm$  18.32}\\ 
    & 15 &{55.03 $\pm$ 9.77} &{57.13 $\pm$ 20.18} 	&{51.42 $\pm$ 27.79}	&{54.71 $\pm$ 15.68}\\
    & $\infty$ (No noise)	&{79.91 $\pm$ 1.87}	&{79.86 $\pm$ 2.98} &{69.73 $\pm$ 2.19 }	&{78.13 $\pm$ 0.76 }\\ 
    \midrule
    \multirow{4}{*}{FL-Non IID}	
    &0.5	&{56.42 $\pm$ 24.13}	&{51.54 $\pm$ 21.57}	&{20.73 $\pm$ 13.67} &{56.05 $\pm$ 17.58}\\ 
    & 5	&{65.36 $\pm$ 3.54}	&{51.54 $\pm$ 17.61}	&{49.32 $\pm$ 16.88}	&{64.24 $\pm$ 7.26}\\ 
    & 15 &{42.91 $\pm$ 16.41} &{42.51 $\pm$ 17.69} 	&{54.24 $\pm$ 17.29 }	&{59.23 $\pm$ 13.10}\\
    & $\infty$ (No noise)	&{73.72 $\pm$ 2.32}	&{74.58 $\pm$ 1.65} &{69.25 $\pm$ 2.04}	&{74.25 $\pm$ 1.78}\\ 
    \bottomrule
  \end{tabularx}
  
  \label{tab:dep}
\end{table*}

\vspace{-2ex}
\subsection{Differential Privacy (DP)}

Differential privacy is an approach that guarantees users not to be affected, adversely or otherwise, by allowing their data to be used in any analysis \cite{DworkRoth14}. It presents strong confidentiality in statistical databases and machine learning approaches through mathematical definition which is an acceptable measure of privacy protection \cite{Dwork08}.

\textbf{Definition 1.1 : }\textit{$M$ and $S$ denote a random mechanism and each output respectively. $D$ and $D'$ are defined neighboring datasets having difference with one record. ($\varepsilon$, $\delta$) protects confidentiality \cite{Dwork11}.}

\begin{equation}
\Pr\left[M\left(D\right) \in S\right] \leq e^{\varepsilon }\cdot \Pr\left[M\left(D'\right) \in S\right] + \delta
\end{equation}

where $\varepsilon$ represents the privacy budget and $\delta$ symbolizes the probability of error. Privacy budget provides controlling the privacy guarantee level of $M$ \cite{Haeberlen11}. The ratio between the two mechanisms ($M(D)$ and $M(D’)$) constraints by $e^\varepsilon$. Where $\delta$ $=$ $0$, M gives $\varepsilon$-differential privacy by its strictest definition. In other case, for some low probability cases, ($\varepsilon$,$\delta$)-differential privacy ensures latitude to invade strict $\varepsilon$-differential privacy. $\varepsilon$-differential privacy is called as pure differential privacy and ($\varepsilon$, $\delta$)-differential privacy, where $\delta$ $>$ $0$, is called as \textit{approximate differential privacy} \cite{beimel2014private}. It is possible to implement differential privacy in two settings: Centralized DP (CDP) and Local DP (LDP) \cite{qu2021privacyadaptive}.

\vspace{-2ex}
\subsection{Federated Learning (FL)}

Since conventional centralized learning systems require that all training data produced on different devices be uploaded to a server or cloud for training, it may give rise to serious privacy concerns \cite{appledp17}.
FL allows training an algorithm in a decentralized way \cite{mcmahan2017communicationefficient, McMahan2016FederatedLO}. It ensures multiple parties collectively train a machine learning model without exchanging the local data \cite{li2021survey}. To define mathematically, it is assumed that there are $N$ parties, and each party is showed with $T_i$, where $i$ $\in$ $[1,N]$. For the non-federated setting, each party uses its local data and depicted by $D_i$ to train a local model $M_i$ and send the local model parameters to the server. The predictive data is sent only the local model parameters to the FL server. 

\begin{table*}[h!]
\footnotesize
\centering
  \caption{Average Test accuracies of models trained in centralized and FL setups, on Sexual Harassment dataset}
  \begin{tabularx}{{0.76\textwidth}}{@{}lrrrrr@{}}
  \toprule
  \textbf{Setup}	&\textbf{Epsilon}	&\textbf{BERT}	&\textbf{RoBERTa}	&\textbf{DistillBERT}	&\textbf{ALBERT}\\
  \midrule 
    \multirow{4}{*}{Centralized DP}
    & 0.5	&{47.89 $\pm$ 4.44} &{38.79 $\pm$ 15.38}	&{38.48 $\pm$ 25.96}	&{48.35 $\pm$ 5.45}\\ 
    & 5	&{51.13 $\pm$ 4.90}	&{42.77 $\pm$ 2.90}	&{36.86 $\pm$ 12.38}	&{51.08 $\pm$ 4.90}\\ 
    & 15	&{51.27 $\pm$ 4.80} &{48.45 $\pm$ 5.36} 	&{47.06 $\pm$ 2.96}	&{54.97 $\pm$ 0.57}\\
    & $\infty$ (No noise)	&{83.56 $\pm$ 0.00}	&{81.14 $\pm$ 0.00} &{73.65 $\pm$ 0.00}	&{56.16 $\pm$ 0.00}\\ 
    \midrule
    \multirow{4}{*}{FL-IID}
    &0.5	&{47.43 $\pm$ 2.94}	&{31.38 $\pm$ 11.66}	&{48.35 $\pm$ 4.45} &{45.26 $\pm$ 0.12}\\
    & 5	&{51.38 $\pm$ 4.22}	&{48.45 $\pm$ 4.38}	&{36.47$\pm$ 19.22}	&{46.46 $\pm$ 1.56}\\ 
    & 15	&{41.49 $\pm$ 12.44} &{60.55 $\pm$ 11.05} 	&{47.31 $\pm$ 3.65}	&{49.19 $\pm$ 2.50}\\ 
    & $\infty$ (No noise) &{71.52 $\pm$ 2.03}	&{76.28 $\pm$ 0.11} &{65.77 $\pm$ 0.85}	&{72.44 $\pm$ 1.74}\\ 
    \midrule
    \multirow{4}{*}{FL-Non IID}	
    &0.5  &{51.13 $\pm$ 3.98}	&{52.33 $\pm$ 1.63} &{15.67 $\pm$ 19.02}	&{47.15 $\pm$ 2.55}\\ 
    & 5	&{48.45 $\pm$ 4.18}	&{50.25 $\pm$ 3.46} &{33.43 $\pm$ 10.92}	&{51.41 $\pm$ 4.19}\\ 
    & 15	&{50.15 $\pm$ 3.91}	&{48.46 $\pm$ 4.37}  &{45.44 $\pm$ 2.73} &{48.95 $\pm$ 5.20}\\
    & $\infty$ (No noise)	&{63.52 $\pm$ 2.25}	&{73.55 $\pm$ 2.95} &{60.47 $\pm$ 2.04}	&{64.21 $\pm$ 8.15}\\ 
    \midrule
  \end{tabularx}
  
  \label{tab:harassment}
\end{table*}


\vspace{-2ex}
\section{Experimental Results}

Two types of data sets, which include tweets to detect depression tendency (called as Depression Dataset in the paper) and sexual harassment, were trained on four NLP models mentioned above by implementing DP and FL in the study. The datasets are splitted into train set and test set with 0.8 train size as shown in Table~\ref{tab:trainingdatasets}. 
Both of the datasets are web scraped from Twitter for the purpose of this study and data cleansing was performed through scripting. 

\begin{table}[h!]
\centering
\caption{Dataset Specifications}
\label{tab:my-table}
\begin{tabularx}{{0.85\textwidth}}{lllll}
\cline{2-4}
\multicolumn{1}{l|}{} &\multicolumn{1}{l|}{Train Split} &\multicolumn{1}{l|}{Train Size} & \multicolumn{1}{l|}{Test Size} &  \\ \cline{1-4}
\multicolumn{1}{|l|}{Depression}        &\multicolumn{1}{l|}{0.8}        &\multicolumn{1}{l|}{2477}      &\multicolumn{1}{l|}{619}      &  \\ \cline{1-4}
\multicolumn{1}{|l|}{Sexual Harrasment} &\multicolumn{1}{l|}{0.8}        &\multicolumn{1}{l|}{2883}      &\multicolumn{1}{l|}{721}      
& \\ 
\cline{1-4}
\end{tabularx}
\label{tab:trainingdatasets}
\vspace{-2ex}
\end{table}

In this section we discuss the results presented in the tables. It should be noted that the tables contain the average and the standard deviation of the results obtained after running the models thrice.
Table~\ref{tab:dep} and ~\ref{tab:harassment} show a comparison according to epsilon values between four language models using Centralized DP. We utilize~\citeauthor{Opacus} for our experiments. We implement DP, FL and DP-FL on BERT, RoBERTa, DistillBERT and ALBERT for $\epsilon=0.5,5,15,\infty$. 


It is presented a benchmark between test accuracies of four language models by implementing DP on Depression Dataset In Table 4. Experiments include the results for different epsilon values and baseline form. On baseline mode, RoBERTa model shows performance in terms of test accuracy value as it has the highest model parameters. But when we compare performance loss in terms of test accuracy decrease, ALBERT shows better performance comparing with other models. In general, we also notice that with the increase in epsilon values, the amount of standard deviation decreases as the model approaches towards its vanilla variant (without DP noise).

Table~\ref{tab:dep} and Table~\ref{tab:harassment} also show the results obtained when only FL \footnote{Note that DP-FL with $\epsilon=\infty$ would correspond to the FL variant.} and DP with FL are applied together (DP-FL) for both IID (data distributed uniformly) and Non-IID data silos. For Non-IID scenarios, we assume $10$ shards of size $240$ assigned to each client. We run it over $10$ clients in total, selecting only a fraction of $0.5$ in each round for training. We add DP locally, that is, to each client model at every iteration and aggregate them to perform Federated Averaging~\cite{mcmahan2017communicationefficient}.
We observe the best accuracies with BERT for the FL implementation and followed closely by RoBERTa, owing to their complex architectures. The results also show that the accuracy decreases by adding DP to FL implementation. When FL and DP are applied independently, it is seen that FL performs better in both data sets as it benefits from different local, client models.



\section{Conclusion \& Future work}
Risks of collecting and sharing individuals' data can limit studies, especially in the healthcare domain.  
Therefore, appropriate privacy measures need to be taken in order to use a person's data, such as the text they write, for diagnostic purposes.
%
In this paper, we compare the utility of central and federated training of BERT-based models, for different levels of privacy ($\varepsilon$ in DP), using depression and sexual harassment-related Tweets.
%
%
%
%
Our empirical studies show that (1) Smaller networks such as ALBERT and DistillBERT degrade much more gracefully than larger models like BERT and RoBERTa, when differentially private training is employed. (2) In the Non-IID setup for FL, which is the realistic scenario in medical applications, utility degradation is on average higher than IID setup, which points at the need for training algorithms tailored to such setups. (3) When the training dataset size is small, DP's effect on utility is more detrimental than when more data is provided~\cite{jana2021investigation, tramer2020differentially}. As future work, an ultimate goal is to build a differentially private federated learning setup for classification in medical use-cases, that compromises the utility as little as possible.



\bibliography{dp}
\bibliographystyle{icml2021}

\end{document}